\documentclass[letterpaper, 10 pt, conference]{ieeeconf}
\IEEEoverridecommandlockouts
\usepackage{gensymb}
\usepackage{graphicx}
\usepackage{amsmath}
\usepackage{cite}
\usepackage{xcolor}
\usepackage{xcolor}
\usepackage{url}
\title{\LARGE \bf
Learning Ankle Exoskeleton Control for Reactive Standing Balance Through Neuromusculoskeletal Simulation}
\author{Maria T. Tagliaferri and Inseung Kang
\thanks{M. T. Tagliaferri and I. Kang are with the Department of Mechanical Engineering, Carnegie Mellon University, Pittsburgh, PA 15213 USA (e-mail: {\tt\footnotesize mtagliaf@cmu.edu}).}
}

\begin{document}

\maketitle
\thispagestyle{empty}
\pagestyle{empty}

\begin{abstract}
Deficits in reactive balance can increase fall risk among older adults and individuals with neuromuscular impairments, motivating the need for interventions that improve balance recovery. Wearable robotic exoskeletons offer a promising approach to addressing this need by augmenting balance control. However, developing controllers that deliver appropriately timed and scaled assistance remains challenging, as evaluating each candidate strategy requires repeated exposure to destabilizing perturbations. We propose a simulation-to-real framework that combines motion imitation and reinforcement learning to develop balance-assistive ankle exoskeleton control. A muscle-driven neuromusculoskeletal model was trained using a small set  of motion-capture data from a single subject to reproduce kinematic and kinetic responses to backward-translation perturbations. This model served as a virtual human for training a reinforcement-learning exoskeleton controller. The resulting controller generated bilateral ankle assistance using only inertial sensor measurements, without explicit perturbation detection. We implemented the controller on-board a bilateral ankle exoskeleton and evaluated it in five healthy subjects whose data were not used for training. Assistance increased the perturbation magnitude that subjects could withstand without stepping by 11.8\% and reduced peak center-of-mass displacement and velocity by 13.5\% and 28.2\%, respectively. Individual benefits varied with baseline recovery strategy and body mass. These results indicate that a single virtual human can yield transferable balance assistance, and that more diverse virtual humans could further improve its effectiveness.

\end{abstract}

\begin{keywords}
Postural Stability, Neuromuscular Simulation, Exoskeleton 
\end{keywords}

\section{Introduction}
Maintaining upright standing balance requires continuous neuromuscular control to regulate the body's center of mass (CoM), which is positioned above a relatively small base of support (BoS). This control relies on coordinated muscle activity across the ankle, knee, and hip joints to regulate CoM motion and maintain postural stability. Impairments in these mechanisms can compromise balance and increase susceptibility to falls, which are a major source of injury among older adults and individuals with neuromuscular impairments \cite{10.1093/ageing/afp200,Wolfson2001}. Wearable robotic exoskeletons offer a potential approach to augment these balance responses by supplementing the joint moments involved in CoM regulation. Several studies have shown that assistive exoskeleton torque can improve measures of postural stability following a balance perturbation \cite{doi:10.1126/scirobotics.adf1080,Emmens2018,10719782}.

Existing balance-assistive controllers generally apply a predefined torque profile during perturbation. For example, studies have evaluated brief plantarflexion torque pulses at the ankle during standing perturbations \cite{doi:10.1126/scirobotics.adf1080,KIM2026113574} or spline-based hip torque profiles during walking perturbations \cite{Monaco2017}. Although these approaches demonstrated the benefits of assistive torque, the assistance was not adaptive to the user's real-time balance response. An alternative approach is to leverage the underlying mechanisms of human balance control to inform exoskeleton assistance. Jakubowski et al. demonstrated that physiologically inspired CoM feedback models can predict coordinated, multi-joint balance-correcting torques across different perturbation conditions and users \cite{Jakubowski20262}. These findings suggest that biological balance-control strategies could provide a basis for developing adaptive, physiologically grounded assistance. However, whether reproducing these torque patterns through an exoskeleton improves balance recovery remains unclear. Furthermore, implementing CoM-based feedback on wearable devices requires accurate, real-time estimation of whole-body CoM motion.

Therefore, determining the appropriate timing and magnitude of assistance to improve postural stability during perturbation recovery remains a central challenge \cite{Tagliaferri2026,KIM2026113574}. Beck et al. demonstrated that ankle assistance improved standing balance only when delivered before the onset of physiological reactive joint moments \cite{doi:10.1126/scirobotics.adf1080}. Regarding assistance magnitude, moderate ankle torque consistently improved CoM displacement during standing perturbations, whereas higher torque magnitudes produced greater inter-subject variability without additional group-level benefits \cite{KIM2026113574}. These findings indicate that multiple assistance profiles must be evaluated to identify effective balance assistance across users. Systematic parameter exploration and human-in-the-loop optimization have identified effective assistance profiles for reducing the metabolic cost of locomotion \cite{Franks2021, 10858452,10.3389/fbioe.2023.1006326}, but both require repeated human-subject testing across conditions. For reactive balance, each evaluation also requires exposure to destabilizing perturbations, which increases experimental burden and concerns regarding subject fatigue and habituation \cite{Tagliaferri2026,Zaback2021}.

Physics-based neuromusculoskeletal simulation offers an alternative for developing and evaluating balance-assistive controllers before human-subject testing. Simulation-to-real transfer of exoskeleton controllers has already been demonstrated for locomotor tasks, such as walking, with an emphasis on improving energetic outcomes \cite{Luo2023,Leem2026,Park2026}. Separately, neuromusculoskeletal simulations have been used to investigate human neuromuscular control and postural stability during perturbed standing \cite{Jones2024,Akbas2025}. However, these two areas have not yet been combined to develop and deploy exoskeleton controllers specifically for reactive balance. By reproducing human responses to perturbations, neuromusculoskeletal simulation provides a controlled environment for learning exoskeleton assistance policies using stability-focused objectives. This approach reduces the number of human-subject perturbation trials needed during controller development, which reduces experimental burden and habituation. Human-subject experiments are then reserved for validating the transferred controller.

In this work, we present a simulation-to-real framework for learning ankle exoskeleton assistance for reactive standing balance. Using approximately 2 minutes of motion-capture data from a single participant, we first train a muscle-driven neuromusculoskeletal model to reproduce human responses to backward-translation perturbations. We then use this model as a virtual human to train a separate reinforcement-learning controller that learns stability-oriented ankle assistance using only device-available sensor inputs, without explicit perturbation detection. Finally, we deploy the learned controller on a bilateral ankle exoskeleton to evaluate its real-time performance and balance benefits in five participants whose data were not used for training.

\section{Methods}
\subsection{Musculoskeletal Simulation}
Simulations were conducted using the Hyfydy physics engine integrated with the SCONE framework \cite{Geijtenbeek2019,Geijtenbeek2021}. Our framework consisted of two sequential training stages: first training a human policy to reproduce recorded balance responses, then training an exoskeleton policy using the simulated human model.

\subsubsection{Reference Motion}
To train the human policy, we collected 24 motion-capture trials (5 s each) from a single subject (height: 160 cm; body mass: 57 kg). During each trial, the subject experienced a backward-translation treadmill perturbation while standing. Perturbation magnitudes ranged from 2 cm to the subject's balance threshold (20 cm), defined as the belt displacement at which the subject lost balance 50\% of the time \cite{palmer2021cortical}. Each magnitude was repeated twice to capture response variability. Trials in which the subject stepped or used their arms to maintain balance were excluded from the training dataset. Reference joint angles and moments were calculated using inverse kinematics and inverse dynamics in OpenSim with a subject-specific, scaled musculoskeletal model \cite{4352056}. Body-segment positions were calculated separately using the unscaled model to map the experimental motion to the generic musculoskeletal model used in simulation.

\begin{figure}[t]
\centering
\includegraphics[width=1\linewidth]{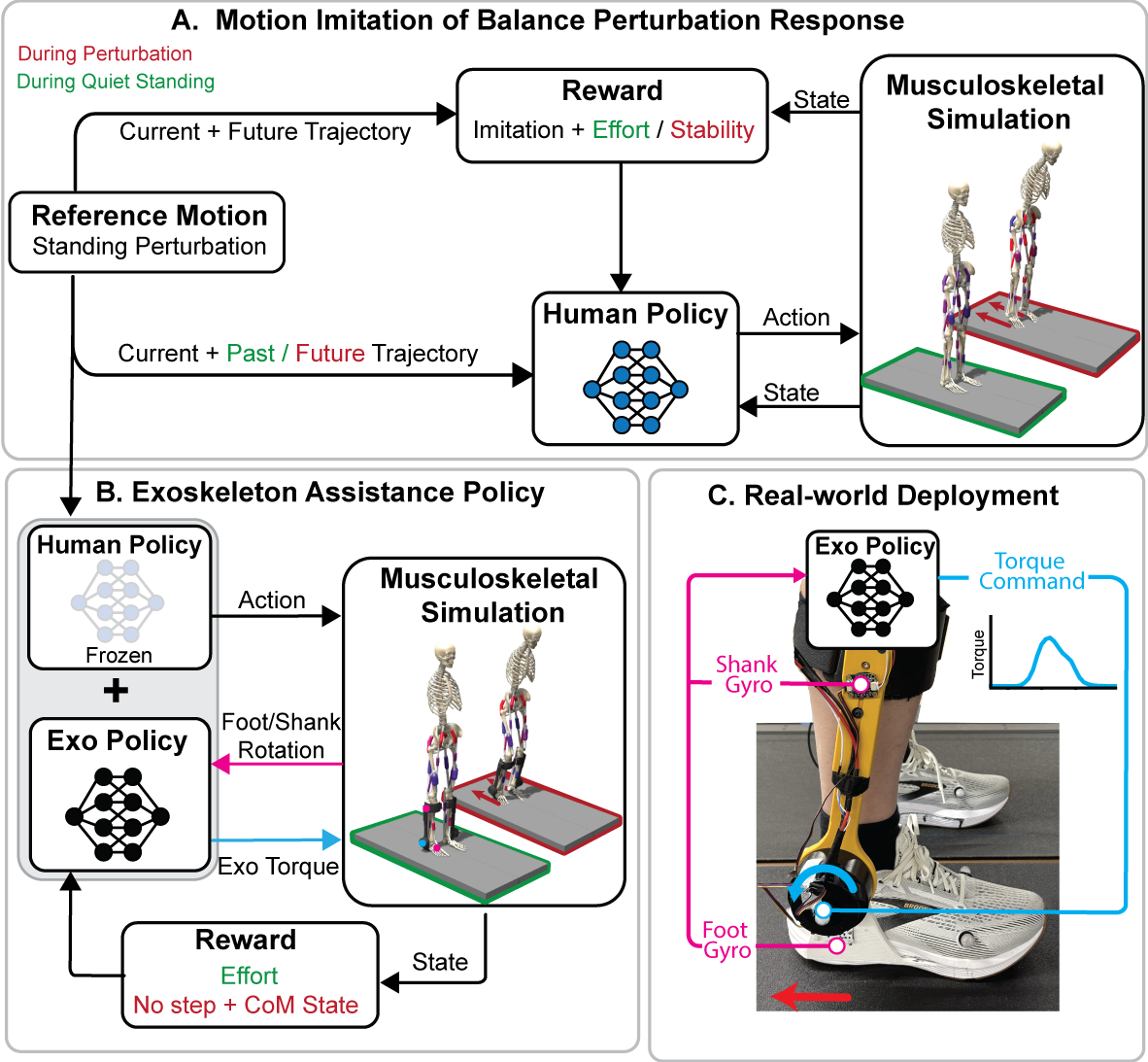}
\caption{Overview of the simulation-to-real framework. (A) Stage 1: training the human policy to reproduce recorded responses to backward-translation perturbations. (B) Stage 2: training the exoskeleton controller to reduce CoM excursion and increase the balance threshold of the simulated human. (C) Deployment of the trained balance-assistive controller on an ankle exoskeleton. The policy receives on-board sensor inputs and outputs a torque command.}
\label{fig:pol}
\end{figure}

\subsubsection{Human Model}
We used a custom 2D musculoskeletal model (modified from H0918, mass: 74.53 kg, height: 1.68 m) \cite{Geijtenbeek2021} with 10 degrees of freedom and 18 muscles. To allow lumbar actuation, we substituted the passive torso joint with an ideal joint motor. A functional split-belt treadmill was modeled to replicate the experimental belt translations and deliver perturbations aligned with the recorded reference data.

\subsubsection{Human Policy Training}
We trained the human control policy using the Soft Actor--Critic (SAC) algorithm \cite{Haarnoja2018} implemented in Stable-Baselines3 \cite{Raffin2021} (Fig.~\ref{fig:pol}A). The policy generated 19 control signals: 18 muscle excitation signals and one torque command for the torso actuator. Muscle excitation signals were converted into muscle activations through first-order activation dynamics. 


At each 25~ms control step, the policy received a 523-dimensional observation describing the simulated human's current state. This included muscle states, joint angles and velocities, and ground reaction forces. The observation included reference motion for the next eight control steps (25–200 ms), consisting of recorded joint angles, joint velocities, and foot positions, along with their corresponding tracking errors. The policy received no explicit information about the perturbation onset, duration, or direction.

Following the DeepMimic motion-tracking framework \cite{2018-TOG-deepMimic}, the human policy was trained to track the reference kinematics using exponential reward terms, with additional effort penalties to limit excessive muscle activation. To adapt this framework to perturbation-evoked balance responses, we introduced three modifications. (1) We added joint-moment tracking rewards to capture the recorded kinetic response in addition to kinematics. (2) During early training, perturbations drove the model far from the reference trajectory, where exponential rewards approach zero and provide no gradient. To address this, we introduced additional Huber penalties \cite{Huber1992} during a 1 s post-perturbation recovery window, which provided a nonzero gradient far from the reference and encouraged the model to reproduce the recorded recovery response. (3) We introduced a time-varying effort penalty that was reduced during the post-perturbation recovery window. This allowed the policy to generate the elevated muscle activations required for balance responses while maintaining effort regularization during quiet stance. The human policy was trained for 30 million environment steps, after which its weights were frozen for Stage 2.

\subsubsection{Exoskeleton Model and Policy Training}
An ankle exoskeleton was added to the human model as an ideal torque actuator with additional mass distributed to the shank, foot and pelvis, bilaterally. The SAC actor for Stage 2 comprised the frozen human-policy branch and a trainable exoskeleton branch, implemented as a multilayer perceptron. The exoskeleton branch received a 40-dimensional observation comprising an 8-step (200 ms) history of sagittal angular velocities from the bilateral shanks and feet and the previous exoskeleton torque command. To approximate device sensors, the simulated gyroscope signals included trial-specific bias and noise and were filtered using an 8 Hz low-pass filter. The branch output a single normalized torque command in $[-1,1]$, shared by both ankles. This command was passed through a 0.15 deadband, scaled to a maximum torque of 16 Nm, and filtered at 6 Hz to approximate actuator dynamics.

During training, the frozen human controller continued generating actions from its original observations but received no imitation reward. The critics had access to the full simulated human state, while the exoskeleton actor remained restricted to the device-available sensor inputs. The exoskeleton branch was trained using SAC for 10 million environment steps. Its objective penalized stepping and CoM displacement and velocity during post-perturbation recovery, along with torque magnitude. Torque magnitude penalties were relaxed during recovery to permit corrective assistance while discouraging unnecessary actuation during quiet stance. No target assistance onset, timing, or torque magnitude was prescribed or explicitly rewarded. Instead, assistance timing and magnitude emerged from optimizing the CoM-based recovery objective, while smoothness penalties refined the resulting assistance. Simulated backward-translation perturbations were delivered at random timing and magnitudes centered around the 20 cm balance threshold.

\subsection{Human Experiment}
\begin{figure}
\centering
\includegraphics[width=1\linewidth]{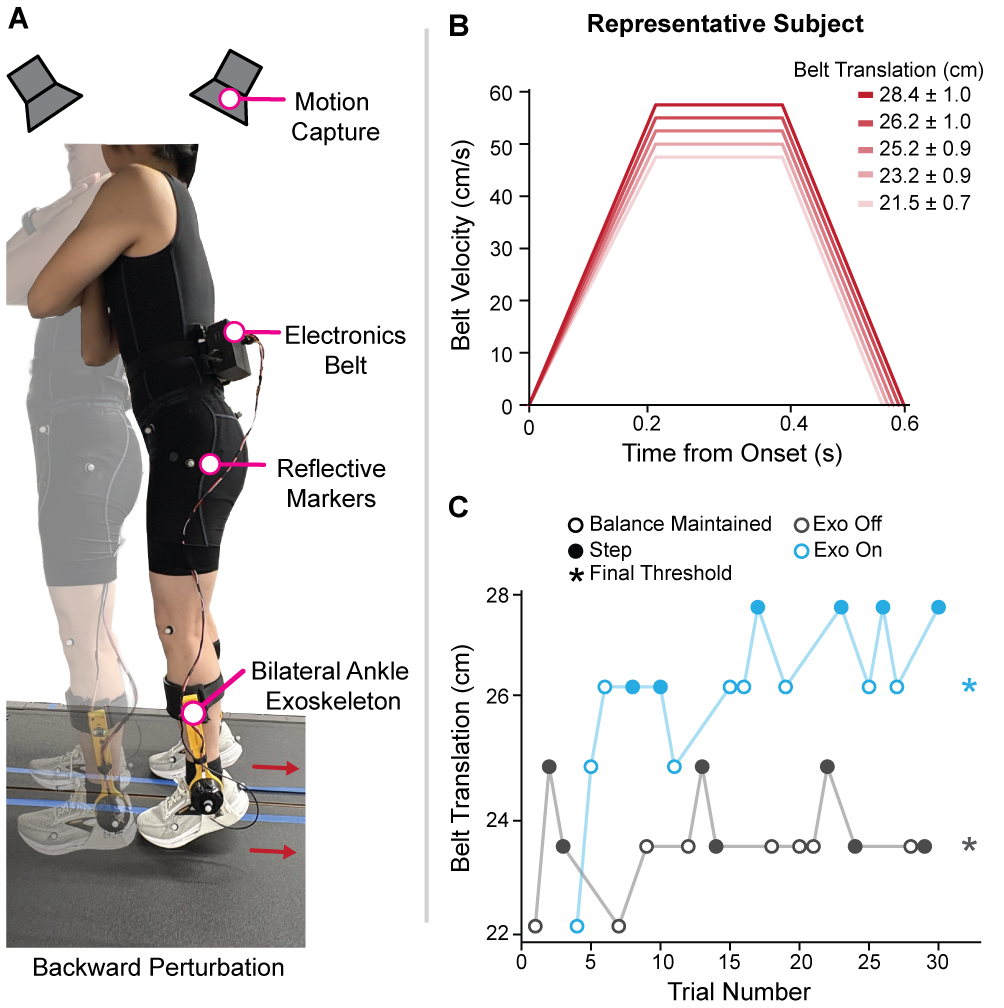}
\caption{Experimental protocol. (A) Participants wore a bilateral ankle exoskeleton while standing on a force-instrumented treadmill and experienced backward-translation perturbations. Foot placement was marked to standardize stance width across trials. (B) Treadmill perturbation profile and target belt displacement for a representative participant. (C) Perturbation-level selection by the QUEST adaptive algorithm and threshold selection for a representative participant.}
\label{fig:exp}
\end{figure}

To validate real-time performance of the exoskeleton policy and test generalization to unseen subjects, we conducted a human subject experiment comparing standing balance with and without the simulation-learned exoskeleton assistance. Five healthy adults (two female; age: 24.4$\pm$0.5 yrs, height: 171.6$\pm$6 cm, body mass: 61.3$\pm$8.7 kg) participated. All participants provided written informed consent and the protocol (STUDY2024\_00000506) was approved by the
Carnegie Mellon University Institutional Review Board.

\subsubsection{Robotic Ankle Exoskeleton}
We developed a bilateral robotic ankle exoskeleton to deliver sagittal-plane torque assistance during standing balance perturbations (Fig.~\ref{fig:exp}A). The device comprises two brushless DC motors with integrated encoders (R02, Robstride, China), a 22.2~V lithium-polymer battery, and 3D-printed user interfaces. Inertial measurement units (BNO055, Adafruit Industries) were mounted bilaterally at the shank and foot. An embedded computing system (Jetson Orin Nano Devkit, NVIDIA), supported by a custom printed circuit board, executed the simulation-trained controller. The device provides 15$^\circ$ of dorsiflexion and 40$^\circ$ of plantarflexion range of motion, with a maximum torque of 16~Nm in either direction. The device weighs 1.1~kg per side, with an additional 1.2~kg for the electronics belt.

\begin{figure*}[h!]
\centering
\includegraphics[width=1\linewidth]{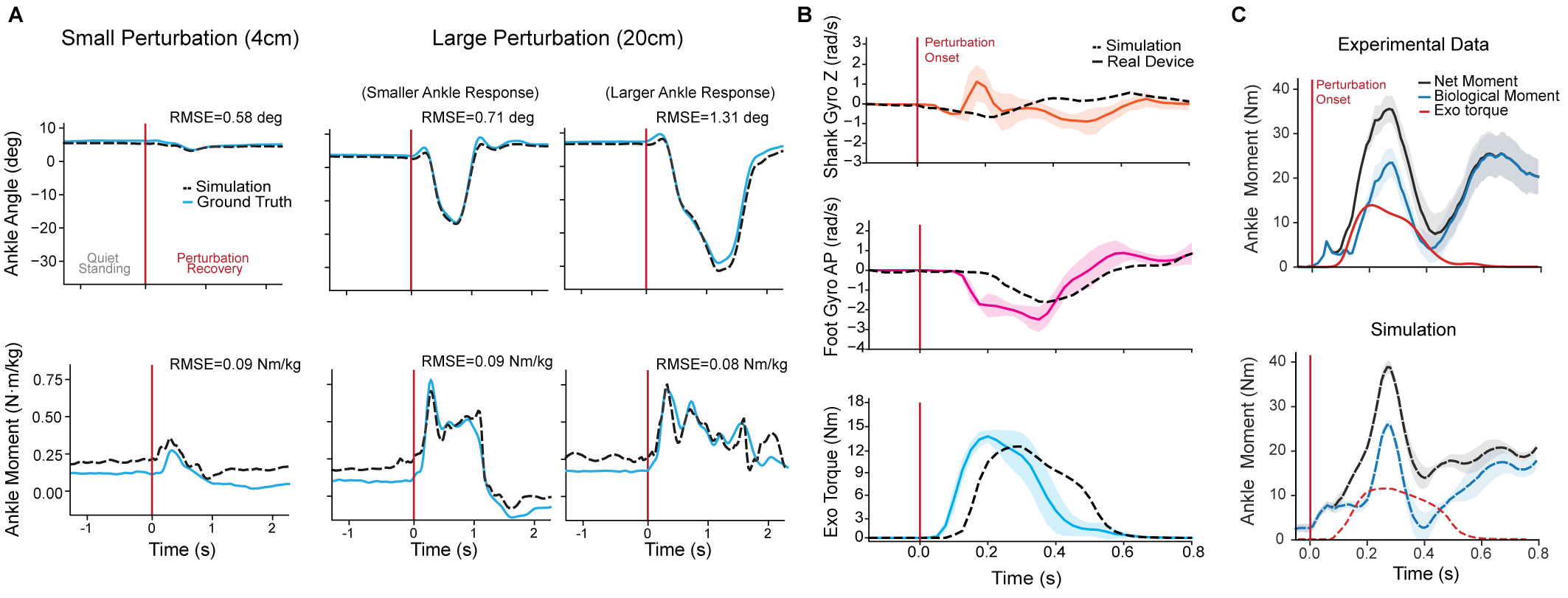}
\caption{Simulation-to-real evaluation of the learned controller. (A) Motion-tracking performance of the Stage 1 human policy in representative trials, showing the response to a small perturbation and response variability across trials of the same larger perturbation. (B) Comparison of simulated gyroscope signals and exoskeleton torque commands with subject-averaged signals recorded during on-device deployment. (C) Comparison of simulated ankle moment offloading with experimentally measured reductions in biological ankle moment during exoskeleton assistance. Red vertical lines indicate onset of the perturbation.  Shaded bands represent $\pm$1 standard deviation. }
\label{fig:imi}
\end{figure*}

\subsubsection{Perturbation Protocol}
Subjects stood on a force-instrumented split-belt treadmill (FIT5, Bertec, USA) with their feet approximately hip-width apart, while wearing the ankle exoskeleton. Foot placement was marked to maintain a consistent stance across trials. Subjects were instructed to cross their arms over their chest and recover balance without stepping (Fig.~\ref{fig:exp}A). Perturbations were delivered by translating both treadmill belts backward simultaneously at a set velocity, acceleration, and time duration \cite{Barnes2023}, producing a target displacement (Fig.~\ref{fig:exp}B). Perturbation onset was randomized to limit anticipation, and unassisted forward perturbations were interspersed to reduce habituation. 

We estimated each subject's balance threshold using the adaptive QUEST algorithm \cite{Watson1983}. QUEST selected each perturbation from 20 predefined belt displacements, adjusting the displacement across trials based on the subject's previous responses (Fig.~\ref{fig:exp}C). A trial was considered successful if the subject maintained balance without stepping or uncrossing their arms. These binary outcomes updated a Bayesian estimate of the balance threshold using a cumulative-normal psychometric function with fixed slope and lapse rate. QUEST was run independently for the exoskeleton-assisted (exo-on) and unassisted (exo-off) conditions, with 15 trials per condition (30 total) interleaved in randomized order. The balance threshold was defined as the belt displacement at which the probability of an unsuccessful recovery was 50\%. Following data collection, we refit the threshold estimates using the complete trial histories for each condition to improve estimation precision. An additional five exoskeleton-assisted trials were collected at each participant's exo-off balance threshold to enable post hoc analysis of balance metrics.

\subsubsection{Data Processing}
Whole-body kinematics were recorded using a motion capture system (Vicon, UK), with reflective markers placed on anatomical landmarks of the torso, legs, and feet. Ground reaction forces were measured using the force-instrumented treadmill. An additional marker on the treadmill belt measured belt displacement and identified perturbation onset. Marker trajectories and ground reaction forces were filtered using fourth-order low-pass filters (6 Hz) and processed in OpenSim \cite{4352056} using musculoskeletal models scaled to each subject. Inverse kinematics and inverse dynamics were used to calculate joint kinematics and moments, while CoM mechanics were derived from the scaled OpenSim models \cite{doi:10.1152/jn.00367.2024}.

\section{Results}
\subsection{Biomechanical Simulation Fidelity}
The human policy tracked the recorded kinematics and kinetics with low error (Fig.~\ref{fig:imi}A). Averaged across all perturbation magnitudes, the model tracked recorded joint angles with a root mean square error (RMSE) of 0.6$\pm$0.2$^\circ$. Moment tracking yielded an RMSE of 0.07$\pm$0.04 Nm/kg. Tracking performance was consistent during both quiet standing and perturbation recovery, capturing trial-to-trial variability in response strategy (Fig.~\ref{fig:imi}A). In simulation, the exoskeleton controller learned a plantarflexion assistance strategy that peaked before the ankle moment response was initiated. The simulated torque assistance increased the balance threshold by 4\% and reduced CoM displacement and velocity compared to the unassisted human model. 

\subsection{Sim-to-Real Transfer}
During real-time deployment, the simulation-trained controller delivered consistent plantarflexion assistance across all five participants (Fig.~\ref{fig:imi}B). Assistance began during the perturbation response without an explicit cue or perturbation detection. Compared with simulation, hardware sensor recordings showed earlier onset in foot angular velocity and a brief increase in shank angular velocity during torque delivery. In hardware, these differences were accompanied by a slightly higher peak torque and earlier assistance onset (87$\pm$12 ms after perturbation onset) than in simulation. 

Additionally, the joint human-exoskeleton policy accurately predicted biological moment offloading during perturbation recovery ($R^2$= 0.81 for ankle moment). All participants exhibited reduced biological joint moments when assisted by the exoskeleton, and this offloading behavior was not diminished with habituation.
\begin{figure}
\centering
\includegraphics[width=1\linewidth]{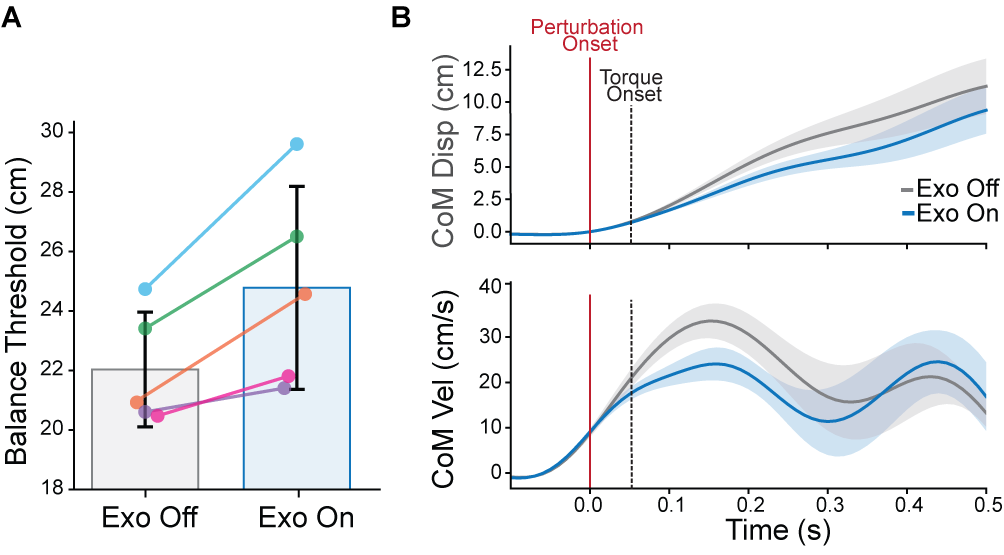}
\caption{Effect of simulation-trained exoskeleton assistance on standing balance outcomes. (A) Comparison of balance thresholds between exoskeleton-assisted (exo-on) and unassisted (exo-off) conditions. Colors represent individual participants and are consistent with those used in Fig.~\ref{fig:tre}. (B) Comparing peak forward CoM displacement and peak CoM velocity (both relative to the BoS) with and without exoskeleton assistance, evaluated at each participant's unassisted balance threshold. Shaded bands represent $\pm$1 standard deviation. }
\label{fig:bal}
\end{figure}
\subsection{Effects of Assistance on Balance Recovery}
The simulation-trained controller increased standing balance threshold compared to the baseline exo-off condition. On average, subjects could withstand 11.8\% higher magnitudes of backward-translation perturbation when receiving torque assistance from the controller (2.8$\pm$1.7 cm threshold increase) (Fig.~\ref{fig:bal}A). 

The learned assistance also reduced CoM excursion (relative to the BoS) compared to the exo-off condition (Fig.~\ref{fig:bal}B). CoM outcomes were calculated from successful trials at each participant's unassisted balance threshold. Exoskeleton assistance reduced peak forward CoM displacement by 13.5$\pm$8.3\% relative to the exo-off condition. The largest reduction in CoM velocity occurred during the first 250 ms after perturbation onset, coinciding with torque delivery, with peak velocity decreasing by 28.2$\pm$5.3\%.

\begin{figure}[!t]
\centering
\includegraphics[width=1\linewidth]{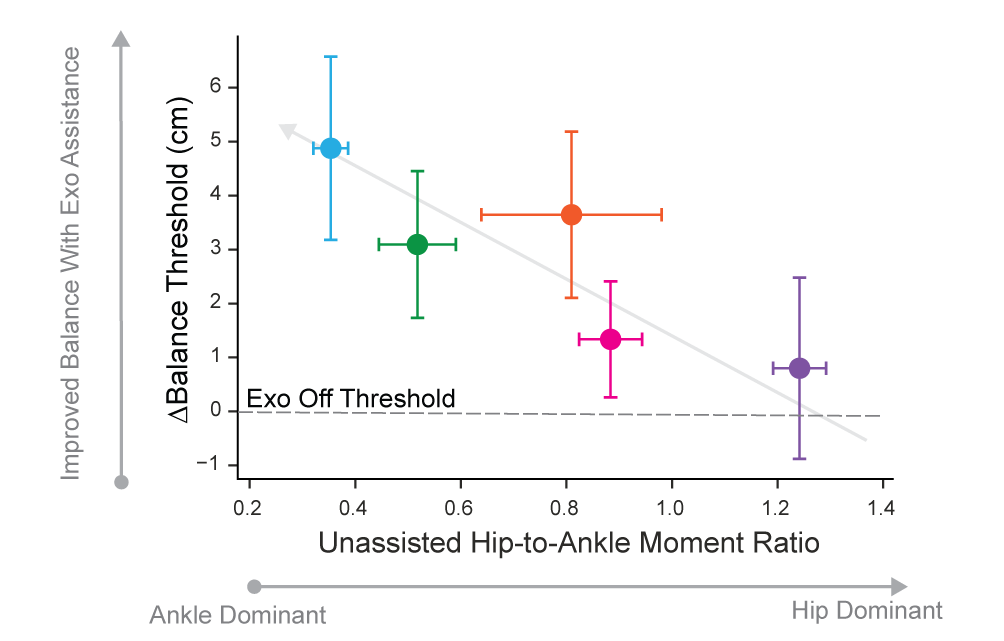}
\caption{Relationship between baseline recovery strategy, quantified by the peak hip-to-ankle moment ratio, and the change in balance threshold with exoskeleton assistance. Colors represent individual participants and are consistent with those used in Fig.~\ref{fig:bal}A.}
\label{fig:tre}
\end{figure}

The effect of the assistance varied across participants (Fig.~\ref{fig:bal}A). Individual improvements in balance threshold were associated with participants' baseline recovery strategies. Participants with more ankle-dominant responses, characterized by higher peak ankle moments relative to hip moments during successful unassisted trials, showed greater improvements in balance threshold (Fig.~\ref{fig:tre}). Threshold gains were also inversely associated with participant body mass, with lighter participants experiencing greater improvements.

\section{Discussion}

To our knowledge, this work is the first to train a balance-assistive ankle exoskeleton controller in neuromusculoskeletal simulation and deploy it on a physical device. A muscle-driven virtual human, trained by motion imitation on approximately 2 minutes of data from a single subject, was sufficient to learn assistance that increased the balance threshold by 11.8\% and reduced peak CoM velocity by 28.2\% in five participants not used for training, without subject-specific tuning. Unlike prior balance-assistive controllers that deliver predefined torque profiles at prescribed delays \cite{KIM2026113574} or rely on a separate perturbation-detection stage \cite{Monaco2017}, the learned policy generated assistance directly from real-time shank and foot angular velocities. Apart from the reference recordings, controller development required no human perturbation trials. Neuromusculoskeletal simulation can therefore facilitate the search for effective assistance timing and magnitude, reserving human-subject experiments for validation. The learned policy delivered plantarflexion torque that peaked before the biological ankle-moment response, consistent with Beck et al.'s findings on the importance of early assistance \cite{doi:10.1126/scirobotics.adf1080}. It also adjusted assistance magnitude and duration to the specific perturbation.

The simulation reproduced key features of the hardware response, including assistance timing relative to the biological response and the reduction in biological ankle moment under assistance (Fig.~\ref{fig:imi}C). However, it underestimated the threshold improvement (4\% in simulation vs. 11.8\% on hardware), and on hardware the assistance began earlier and reached a slightly higher peak torque. Hardware gyroscope signals showed an earlier rise in foot angular velocity and a transient increase in shank angular velocity during torque delivery (Fig.~\ref{fig:imi}B), suggesting that sensor placement and mechanical coupling through the compliant cuffs altered the controller inputs. Although these differences did not prevent the controller from providing sufficient assistance, they highlight that the behavior and performance of simulation-based control depend on the fidelity of the virtual human and the device model.

The benefit of assistance also depended on the user. Participants who relied more on the ankle during exo-off recovery gained the most, suggesting that the effectiveness of ankle assistance depends partly on how individuals naturally regulate balance. The smaller gains in heavier participants may reflect the lower relative assistance provided by the fixed 16 Nm torque limit, which body-mass-normalized torque limits would test directly. The generic virtual human (74.5 kg) also differed in mass from both the reference subject (57 kg) and the participants, so the learned torque magnitude was not matched directly to any individual.

This study has several limitations. First, the virtual human was built from one subject and only from successful trials up to the exo-off balance threshold, so it did not represent the range of anthropometry, recovery strategies, and near-failure responses encountered during hardware testing. Training across populations of virtual humans with varied anthropometry and balance strategies would address this. Second, the frozen human policy could respond to assistance but not adapt to it, and the device model omitted cuff compliance and sensor coupling. Modeling these interactions should narrow the sim-to-real gap in assistance onset and threshold gains. Third, the evaluation included five healthy adults and backward translations only. Larger cohorts, individuals with impaired balance, and additional perturbation directions are needed to establish clinical relevance. Finally, replacing reference-motion imitation with mechanistic models of human balance control would allow the virtual human to generate responses beyond the recorded data, including adaptation to the device.

\section{Conclusion}

This work trained a balance-assistive ankle exoskeleton controller entirely in neuromusculoskeletal simulation and deployed it on hardware without subject-specific tuning. In five participants not used for training, the learned assistance increased the balance threshold and reduced peak CoM displacement and velocity during backward-translation perturbations. The findings of this study motivate extending this framework to more diverse virtual human populations and broader perturbation conditions, so that the learned assistance accounts for the range of body sizes and recovery strategies seen across users.

\bibliographystyle{ieeetr}
\bibliography{Bibliography}
\end{document}